\documentclass[]{TEAI}
\usepackage{helvet}

\usepackage{amsmath} 
\usepackage{natbib}
\usepackage{graphicx}
\usepackage{subcaption} 

\usepackage[toc,page,header]{appendix}
\usepackage[utf8]{inputenc} 
\usepackage[T1]{fontenc}    
\usepackage{hyperref}       
\usepackage{url}            
\usepackage{booktabs}       
\usepackage[nomath]{lmodern} 
\usepackage{amsfonts}       
\usepackage{nicefrac}       
\usepackage{microtype}      
\usepackage{wrapfig}

\usepackage{amssymb}  
\usepackage{fontawesome}  
\usepackage{url}  

\usepackage{titletoc}

\usepackage{tikz}  
\usepackage{comment}  
\usepackage{tabularx}  
\usepackage{booktabs}  

\usepackage{minitoc}

\usepackage{booktabs}
\usepackage{array}
\usepackage{etoolbox}

\definecolor{lightblue}{RGB}{200, 230, 255}  
\definecolor{headerblue}{RGB}{150, 200, 255} 

\usepackage{pgfplots}
\usepackage[utf8]{inputenc} 
\usepackage[T1]{fontenc}    
\usepackage{hyperref}       
\usepackage{url}            
\usepackage{booktabs}       
\usepackage{amsfonts}       
\usepackage{nicefrac}       
\usepackage{microtype}      
\usepackage{xcolor}         
\usepackage{graphicx}
\usepackage{float}
\usepackage{needspace}
\usepackage{comment}
\usepackage{multirow} 
\usepackage{amsmath} 
\usepackage{makecell} 
\usepackage{siunitx}  
\usepackage{tikz}
\usepackage{pgf-pie} 
\usepackage{subcaption}
\usepackage{wrapfig}
\usepackage[export]{adjustbox}

\usepackage{ragged2e}      
\usepackage{tabularx}       
\usepackage{array}          
\usepackage{caption}        
\usepackage{enumitem}
\usepackage{pifont}
\usepackage[hang,flushmargin]{footmisc} 

\usepackage{tcolorbox}

\usepackage{tcolorbox}    
\tcbuselibrary{breakable}  
\tcbuselibrary{skins}      

\usepackage{tabularx}
\usepackage{listings}

\usepackage[table]{xcolor}

\title{More Powerful, Less Verifiable? Mitigating Hard-to-Verify Hallucinations in MLLM's Reinforcement Learning}

\author{
    Jianhong Pang\textsuperscript{1,2},
    Ziyi Ye\textsuperscript{1,2,$\dagger$},
    Qiuxia Hou\textsuperscript{3},
    Yanhao Zhang\textsuperscript{3},
    Zuxuan Wu\textsuperscript{1,2}
}
\affiliation[1]{\mbox{Institute of Trustworthy Embodied AI, Fudan University}} 
\affiliation[2]{\mbox{Shanghai Key Laboratory of Multimodal Embodied AI}} 
\affiliation[3]{\mbox{OPPO AI Center, OPPO Inc.}}

\abstract{
\begin{abstract}
Reinforcement learning (RL) strengthens model capabilities through feedback from humans and environments.
However, emerging evidence suggests that reward optimization can produce outputs whose errors are harder for humans to detect.
To address this issue, we construct \textbf{ITHV} (the \textbf{I}mage-\textbf{T}ext \textbf{H}allucination \textbf{V}erifiability Dataset), comprising 2,728 visually grounded instances 
that distinguish between \textbf{obvious hallucinations}, which humans readily detect, and \textbf{elusive hallucinations}, which humans struggle to identify.
We then propose \textbf{DAPO-V}, an extension of DAPO that maximizes correctness subject to a budget on verifiability-aware risk.
Specifically, DAPO-V combines a verifiability-aware reward with adaptive dual control, where larger penalties are assigned to incorrect responses on questions with lower human verification accuracy.
An adaptive multiplier further adjusts the penalty strength according to risk-budget violations. 
Across three MLLM base models, DAPO-V improves overall accuracy while substantially improving accuracy on elusive hallucination questions by 5.78--9.72 percentage points over accuracy-only DAPO.
We further test DAPO-V with a blinded human study, which shows that hallucinations are more readily detected in DAPO-V than those of the base model.
\end{abstract}
}

\checkdata[Code]{\url{https://github.com/pang-jh/RL-Verifiability}}
\checkdata[Dataset]{\url{https://huggingface.co/datasets/BeEnough/ITHV}}

\begin{document}
\maketitle
\renewcommand{\thefootnote}{}
\footnotetext{$^\dagger$Corresponding authors.}
\footnotetext{Authors' email addresses: Jianhong Pang(\url{jhpang25@m.fudan.edu.cn}), Ziyi Ye(\url{zyye@fudan.edu.cn}).} 
\renewcommand{\thefootnote}{\arabic{footnote}}


\vspace{-1.5em}

\section{Introduction}
\label{sec:intro}

Reinforcement learning (RL) is widely used in language-model post-training for preference alignment and correctness-based optimization~\cite{ouyang2022training,shao2024deepseekmath}.
However, emerging evidence has shown that optimizing these reward objectives may improve model capability while making the remaining errors harder for humans to detect.
For example, \citet{si2024large} show that LLM explanations help users verify claims more efficiently, but can also mislead them when the explanations are incorrect.
\citet{baker2025monitoring} further show that optimizing against a chain-of-thought monitor can improve task performance while teaching agents to conceal reward hacking from the monitor.

As language models become more capable and extend to multimodal scenarios~\cite{qwen2.5-VL,an2025llavaonevision15fullyopenframework}, understanding how RL affects the human verifiability of their outputs becomes increasingly important. 
In image–text settings, users can consult images to verify generated claims, yet some hallucinations remain difficult to detect even with this visual evidence.
For example, describing a visibly cloudy sky as ``clear and cloudless''
creates an obvious hallucination that users can readily reject, whereas describing the boat as ``wooden'' creates an elusive hallucination, as its material is difficult to determine from the image~(see examples in Fig.~\ref{fig:dataset}(b)).

\begin{wrapfigure}{r}{0.5\textwidth}
    \centering
    \includegraphics[width=0.4\textwidth]{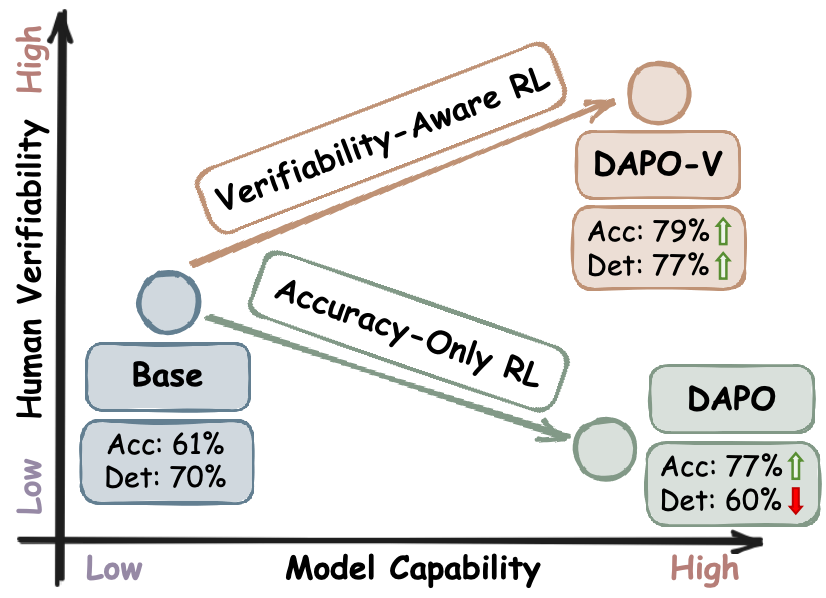}
    \caption{\textbf{RL can induce divergent capability--verifiability paths.}
    We report ITHV accuracy and human detection rates for hallucinations generated by LLaVA-OneVision-1.5-8B.
    }
    \label{fig:motivation}
\end{wrapfigure}

However, addressing the verifiability of hallucinations in MLLMs poses two challenges. 
First, existing datasets provide limited annotations of how easily humans can detect hallucinations, hindering both evaluation of human verifiability and the development of training signals that target errors humans struggle to detect.
Second, effective RL methods that explicitly account for human verification difficulty remain underexplored, limiting our ability to reduce hallucinations that are difficult for humans to identify.
As shown in Fig.~\ref{fig:motivation}, accuracy-only RL may improve model capability while making generated hallucinations harder for humans to detect.
This motivates learning a verifiability-aware RL policy that improves model capability while suppressing errors that are difficult for humans to identify.

To tackle these challenges, this paper investigates this issue in DAPO-trained~\cite{yu2026dapo} multimodal large language models (MLLMs).
Firstly, we construct the \textbf{ITHV} (the \textbf{I}mage-\textbf{T}ext \textbf{H}allucination \textbf{V}erifiability Dataset), comprising 2,728 visually grounded instances across verification, question answering, description, and factual correction tasks. 
ITHV follows prior work that distinguishes \textbf{O}bvious \textbf{H}allucinations (\textbf{OH}), which humans readily identify, from \textbf{E}lusive \textbf{H}allucinations (\textbf{EH}), which are frequently missed or require greater verification effort, and explores their regulation through activation interventions~\cite{pang2026steering}.
ITHV extends \citep{pang2026steering} multiple-choice format to include free-form responses, enabling RL-based investigation of hallucination mitigation under varying human verification difficulty.
Its few-thousand-instance scale is comparable to that used in targeted, data-efficient multimodal RL studies~\cite{yu2024rlhf,huang2026sam}.

Based on the ITHV dataset, we propose \textbf{DAPO-V}, a human-verifiability-aware risk-constrained extension of DAPO~\cite{yu2026dapo}.
\textbf{DAPO-V} uses human accuracy as a question-level proxy for human verifiability and maximizes overall correctness subject to a budget on verifiability-aware risk.
An adaptive multiplier adjusts the penalty according to observed budget violations.
Because DAPO dynamically selects response groups using their correctness patterns, we monitor raw groups before filtering to avoid conditioning risk control on this selection.
We also preserve verifiability-based weighting during advantage computation.
Together, these designs couple DAPO policy updates with human-verifiability-aware risk control.

Experiments on three MLLMs show that DAPO-V improves EH accuracy over accuracy-only DAPO by 5.78--9.72 percentage points and reduces verifiability-aware risk by 10.1--19.5\%, while improving overall accuracy.
A blinded human study further shows that hallucinations generated by DAPO-V are easier for humans to detect and require less verification time.
Finally, we examine whether the verifiability-aware reward also benefits PPO~\cite{schulman2017proximal} and GRPO~\cite{shao2024deepseekmath}.
Their results support the reward mechanism's transferability to other policy optimizers.

In summary, our contributions are as follows:
\begin{itemize}
    \item We construct ITHV, combining visually grounded tasks and human verification metadata to support human-verifiability-aware evaluation and learning.
    \item We develop DAPO-V, incorporating verifiability-aware risk constraints, adaptive dual control, and pre-filtering risk monitoring into DAPO-based policy learning.
    \item We demonstrate improved EH mitigation and lower weighted risk across three models, complemented by blinded human evaluation and additional experiments transferring the reward mechanism to PPO and GRPO.
\end{itemize}

\section{Data Construction}
\label{sec:data_construction}

\begin{figure*}[t]
    \centering
    \includegraphics[width=0.95\textwidth]{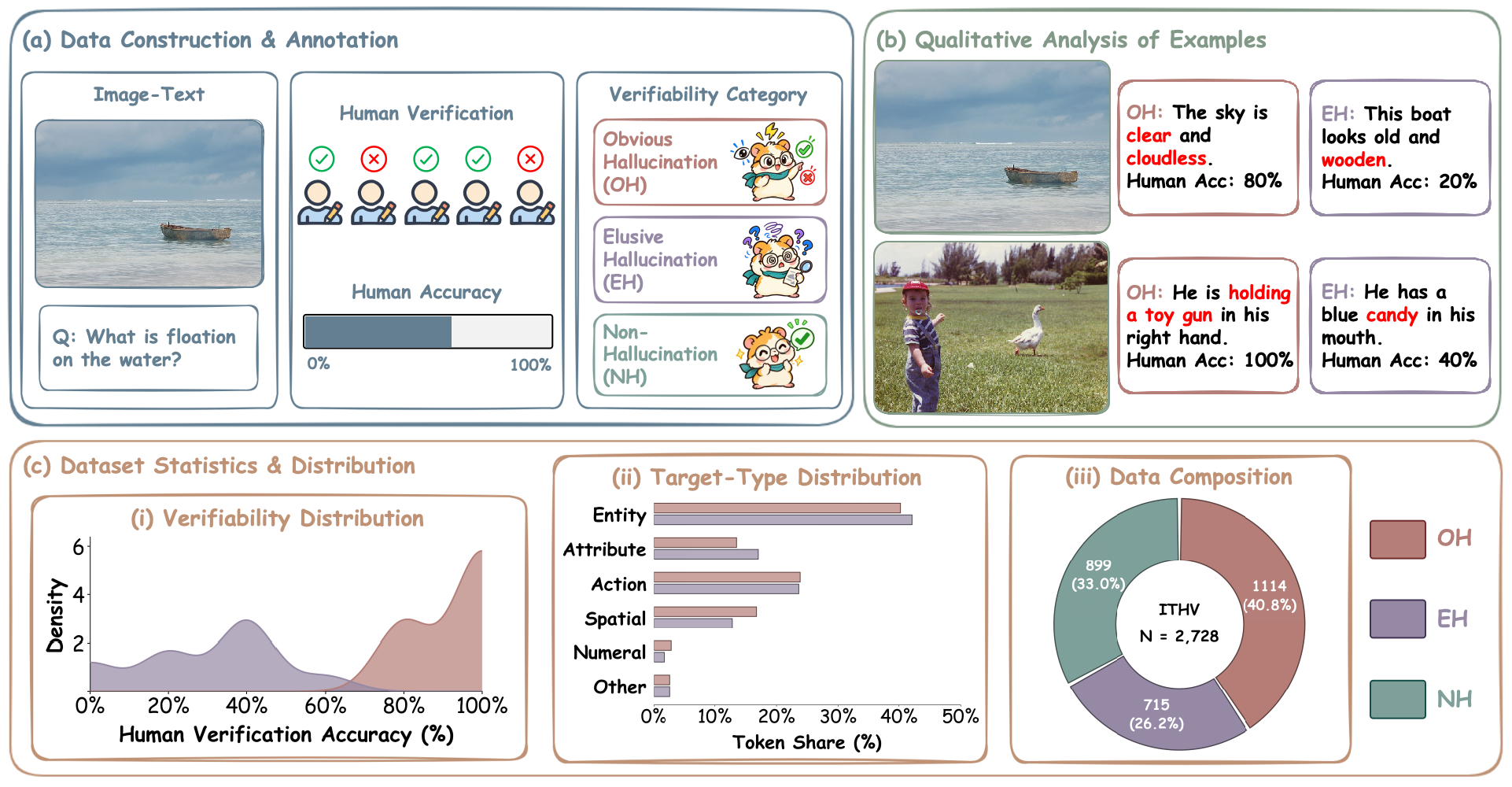}
    \caption{
\textbf{ITHV: construction, qualitative examples, and dataset statistics.}
(a) Data construction and human verification.
(b) Qualitative analysis of representative examples.
(c) Dataset statistics and distributions: (i) human verification accuracy distributions for OH and EH; (ii) verification-target type distributions across the two subsets; (iii) dataset composition.
}
    \label{fig:dataset}
\end{figure*}

We construct the \textbf{ITHV} (the \textbf{I}mage-\textbf{T}ext \textbf{H}allucination \textbf{V}erifiability Dataset) to support human-verifiability-aware evaluation and policy learning.
Correctness labels alone do not capture how difficult an error is for humans to verify.
ITHV therefore combines visually grounded tasks, human verification metadata, and complementary supervision on valid visual content.
Figure~\ref{fig:dataset} summarizes its construction and composition.

\subsection{Task Construction}
\label{sec:ITHV_tasks}
ITHV covers image-grounded binary verification, multiple-choice, short-answer, and factual correction tasks.
Questions and reference answers are manually reviewed for visual grounding and task consistency.
Image sources and construction details are provided in Appendix~\ref{app:ithv_sources}.

\subsection{Human Verification and Categorization}
\label{sec:ITHV_annotation}

We follow the human-verifiability annotation protocol and categorization criteria established in~\cite{pang2026steering}, using human accuracy and response time to distinguish obvious from elusive hallucinations.
Image--text consistency determines whether a claim is hallucinated, whereas human verification results indicate how readily that hallucination can be detected.
The study involved 65 volunteers, with each source item assessed by five independent participants under a 20-second time limit.
Further details about participants and the annotation protocol are provided in Appendix~\ref{app:dataset_participants}.
Among hallucination-bearing items, those with human accuracy $a_i\geq0.8$ are categorized as \textbf{O}bvious \textbf{H}allucinations (\textbf{OH}).
Items with $a_i\leq0.4$, or $a_i=0.6$ and a median response time exceeding 12 seconds, are categorized as \textbf{E}lusive \textbf{H}allucinations (\textbf{EH}).
Thus, EH captures both frequently missed inconsistencies and those requiring greater verification effort.
Items satisfying neither criterion are not assigned to either category.
We retain $a_i$ as a proxy for human verifiability in risk assessment.

\subsection{Dataset Composition and Statistics}
\label{sec:ITHV_statistics}

ITHV contains 2,728 instances: 1,114 OH, 715 EH, and 899 Non-Hallucinated (NH) examples. 
NH instances contain visually supported information and provide complementary supervision intended to discourage indiscriminate rejection and unsupported correction during hallucination-focused training.

Figure~\ref{fig:dataset}(c)(i) presents the human verification accuracy distributions.
Consistent with the categorization criteria, OH instances concentrate at higher accuracy levels, whereas EH instances exhibit lower accuracy, with variation within each category.
We further categorize each sample's minimal verification target into six semantic types.
Figure~\ref{fig:dataset}(c)(ii) shows highly similar distributions across the two subsets, indicating that the verifiability labels reflect human verification outcomes rather than expression types. 
Detailed vocabulary visualizations are provided in Appendix Figure~\ref{fig:english_wordcloud}.
Moreover, Figure~\ref{fig:dataset}(b) illustrates that human verification accuracy can vary substantially across claims associated with the same image, motivating the use of sample-level accuracy as a continuous proxy for human verifiability.

\section{Method}
\label{sec:method}

\begin{figure*}[t]
    \centering
    \includegraphics[width=0.95\textwidth]{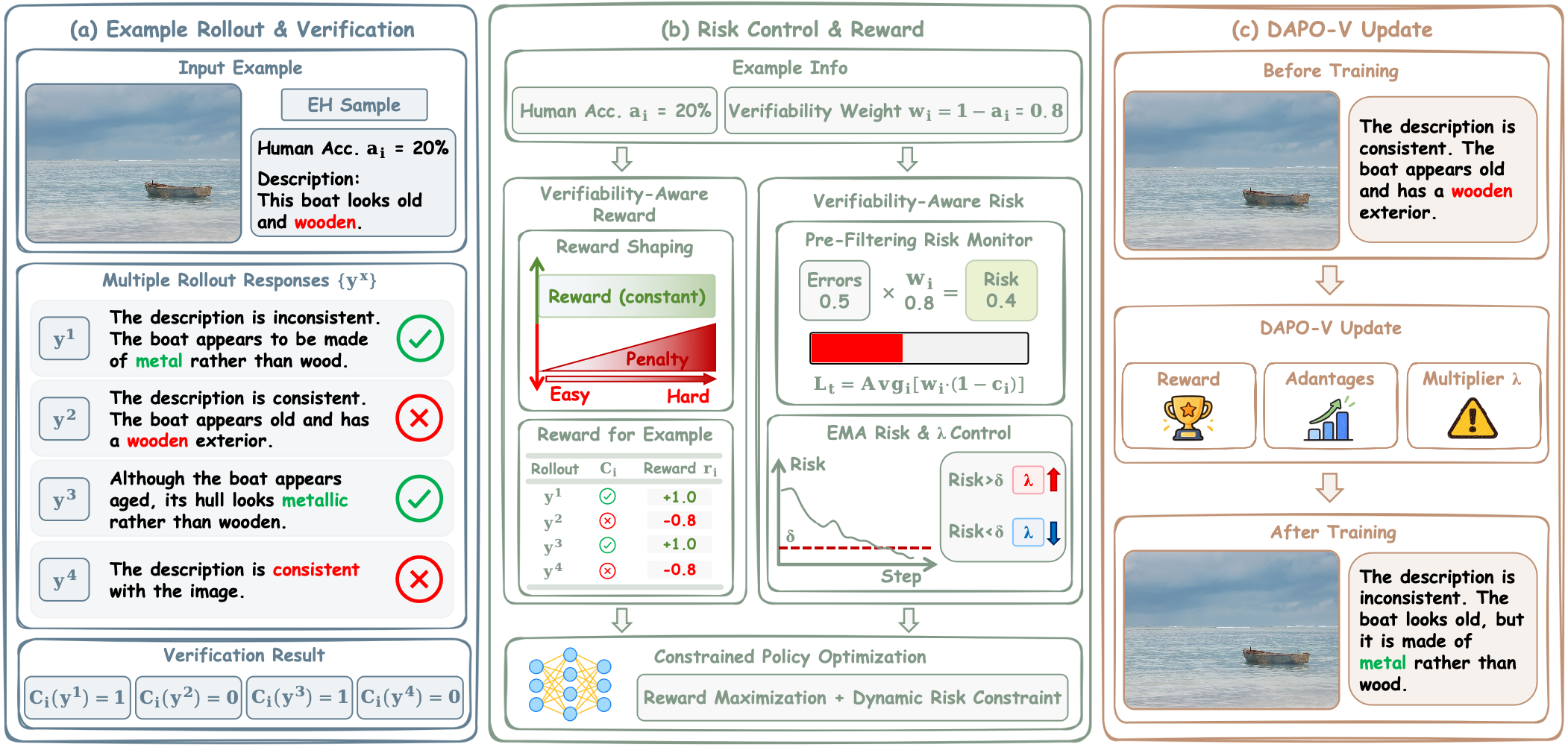}
    \caption{
\textbf{Overview of DAPO-V: human-verifiability-aware
risk-constrained reinforcement learning.}
(a) The policy generates grouped responses.
(b) Human verification accuracy determines the verifiability weight. Correct responses receive constant rewards, while incorrect ones receive verifiability-aware penalties. EMA-smoothed risk guides multiplier updates under a risk budget.
(c) Rewards, advantages, and risk constraints guide policy optimization, illustrated by before-and-after outputs.
}
    \label{fig:framework}
\end{figure*}

We formulate human-verifiability-aware hallucination mitigation as risk-constrained policy learning. 
Our objective is to improve overall answer correctness while limiting errors on questions that humans find difficult to verify. 
Figure~\ref{fig:framework} presents an overview of DAPO-V, which combines a verifiability-aware risk model, adaptive Lagrangian control, and DAPO-based~\cite{yu2026dapo} policy updates.

\subsection{Human Verifiability-Aware Risk Modeling}
\label{sec:risk_modeling}

Let $\mathcal D=\{(x_i,y_i^\star,a_i)\}_{i=1}^{N}$ denote the training set, where $x_i=(I_i,q_i)$ is an image--question pair, $y_i^\star$ is the reference answer, and $a_i\in[0,1]$ denotes empirical human accuracy when available. 
Given $x_i$, the policy $\pi_\theta$ generates a response $y$, whose correctness is evaluated against the reference answer to obtain a binary signal $C_i(y)\in\{0,1\}$.

Let $\mathcal H=\{i:a_i\text{ is available}\}$ denote the annotated subset.
For $i\in\mathcal H$, we use $w_i=1-a_i$ as a proxy for human verification difficulty, assigning larger weights to instances with lower human accuracy.
With $N_H=|\mathcal H|>0$, we define the verifiability-aware risk as
\begin{equation}
L_H(\theta)
=
\frac{1}{N_H}
\sum_{i\in\mathcal H}
w_i\,\mathbb E_{y\sim\pi_\theta(\cdot\mid x_i)}[1-C_i(y)].
\label{eq:risk}
\end{equation}
This formulation jointly accounts for model correctness and human verification difficulty, assigning greater risk to incorrect predictions on harder-to-verify instances.

\subsection{Risk-Constrained Learning Objective}
\label{sec:risk_objective}

We maximize correctness over the full training population while imposing a budget on the risk in Eq.~\eqref{eq:risk}:
\begin{equation}
\max_\theta \quad
J_C(\theta)
=
\frac{1}{N}
\sum_{i=1}^{N}
\mathbb E_{y\sim\pi_\theta(\cdot\mid x_i)}
[C_i(y)]
\qquad
\mathrm{s.t.}\quad
L_H(\theta)\leq\delta.
\label{eq:learning_objective}
\end{equation}
where $\delta\geq0$ specifies the allowable proxy risk.
The correctness objective includes unannotated instances, whereas the constraint is averaged over $\mathcal H$.
This formulation expresses the desired risk level through a budget while retaining correctness as the common learning objective.

\subsection{Risk Monitoring and Dual Control}
\label{sec:dual_control}

\paragraph{Pre-filtering risk monitor.}
DAPO~\cite{yu2026dapo} selects training groups according to their generated correctness patterns.
Using only these groups to estimate risk would condition monitoring on a policy-dependent selection.
We instead monitor a fixed prefix of raw groups before filtering.

At outer iteration $t$, we sample $G$ responses $\{y_{ij}\}_{j=1}^{G}$ per question using $\pi_{\theta_t}$ and write $C_{ij}=C_i(y_{ij})$.
The first $M$ raw question groups form a monitoring window $\mathcal W_t$, including groups that are subsequently discarded.
The monitoring estimate is
\begin{equation}
\widehat L_t
=
\frac{1}{|\mathcal W_t^H|}
\sum_{i\in\mathcal W_t^H}
w_i\left[\frac{1}{G}\sum_{j=1}^{G}(1-C_{ij})\right],
\label{eq:risk_estimate}
\end{equation}
where $\mathcal W_t^H=\mathcal W_t\cap\mathcal H$.
Subsequent refill groups may contribute to policy learning but do not change this window.
Sampling continues until both the retained training batch and the monitoring window are complete.

\paragraph{Verifiability-aware reward.}
We introduce a nonnegative Lagrangian~\cite{tessler2018reward} multiplier $\lambda\geq0$ and construct
\begin{equation}
\mathcal L(\theta,\lambda)
=
J_C(\theta)-\lambda\bigl(L_H(\theta)-\delta\bigr).
\label{eq:lagrangian}
\end{equation}
The policy maximizes $\mathcal L$, while the multiplier is updated in the minimizing direction.

To account for the populations underlying $J_C$ and $L_H$, let $p_H=N_H/N$ and $c_i(y)=w_i[1-C_i(y)]$.
Under uniform sampling of training indices, $\mathbb E_{i,y}[c_i(y)]=p_H L_H(\theta)$.
Consequently, omitting the response-independent term $\lambda\delta$ gives the response-level reward
\begin{equation}
R_t(x_i,y)
=
C_i(y)
-
\frac{\lambda_t}{p_H}\,
w_i[1-C_i(y)].
\label{eq:reward}
\end{equation}
The factor $p_H^{-1}$ preserves the normalization of the annotated-subset risk when policy learning uses the full dataset. 
Thus, correct responses receive a common reward, while incorrect responses incur verifiability-based costs.

\paragraph{Multiplier update.}
We smooth the monitoring estimate with an exponential moving average:
\begin{equation}
\widetilde L_t
=
\beta\widetilde L_{t-1}
+
(1-\beta)\widehat L_t,
\label{eq:risk_ema}
\end{equation}
initialized with the first available estimate.
After the policy update, we perform one projected multiplier update:
\begin{equation}
\lambda_{t+1}
=
\Pi_{[0,\lambda_{\max}]}
\left[
\lambda_t+
\eta_\lambda(\widetilde L_t-\delta)
\right],
\label{eq:dual_update}
\end{equation}
where $\eta_\lambda>0$, $0\leq\beta<1$, $\lambda_{\max}>0$, and $\Pi$ denotes projection.
This feedback increases the penalty above the budget and relaxes it below the budget.

\subsection{Integration with DAPO}
\label{sec:dapo_integration}

We retain DAPO's dynamic group selection and asymmetrically clipped token-level objective~\cite{yu2026dapo}.
For each response group, let $\bar C_i=G^{-1}\sum_j C_{ij}$.
Groups satisfying $\tau_{\mathrm{low}}<\bar C_i<\tau_{\mathrm{high}}$ are retained, where the thresholds exclude groups without correctness variation.
This selection serves policy learning only; the controller uses the pre-filtering window in Section~\ref{sec:dual_control}.

Our integration preserves the verifiability-based weighting of Eq.~\eqref{eq:reward} in the learning signal.
Let $R_{ij}=R_t(x_i,y_{ij})$ and $\bar R_i=G^{-1}\sum_j R_{ij}$.
We then compute the mean-centered relative advantages:
\begin{equation}
\widehat A_{ij}
=
R_{ij}-\bar R_i
=
\left(1+\frac{\lambda_t}{p_H}w_i\right)
(C_{ij}-\bar C_i).
\label{eq:remove_norm}
\end{equation}
To preserve the question-level weight $1+\lambda_t w_i/p_H$ in Eq.~\eqref{eq:remove_norm}, we mean-center group rewards without standard-deviation normalization, which would otherwise cancel this weight up to numerical regularization.
The resulting advantages are used in a DAPO-style policy objective, whose formulation and further implementation details are provided in Appendix~\ref{app:policy_optimization}.
\section{Experiments}
\label{sec:experimental_results}

\subsection{Experimental Setup}
\label{sec:experimental_setup}

\begin{figure*}[t]
    \centering
    \includegraphics[width=0.95\textwidth]{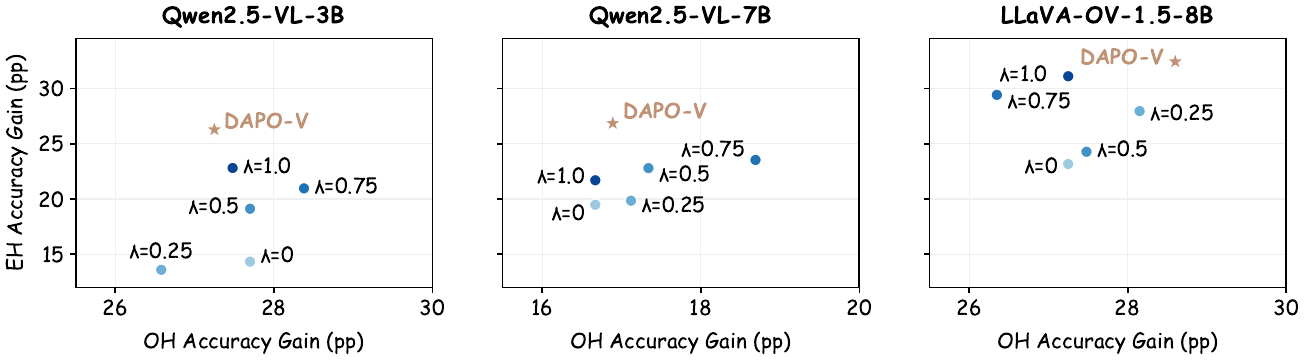}
    \caption{
    \textbf{Accuracy gains on obvious and elusive hallucinations.}
    Each panel reports OH and EH accuracy gains relative to the corresponding base model.
    Circles denote fixed-weight DAPO variants, and stars denote DAPO-V.
    All gains are measured in percentage points.
    Across all three models, DAPO-V achieves the largest EH accuracy gains among the compared settings.
    }
    \label{fig:obvious_vs_elusive}
\end{figure*}

\paragraph{Data and evaluation.}
We conduct experiments on the ITHV dataset and report mean response accuracy overall and for OH and EH separately.
We additionally evaluate the verifiability-aware risk $L_H$
defined in Eq.~\eqref{eq:risk}, estimated from the sampled responses on OH and EH examples.
To characterize differential improvements, we report
\begin{equation}
S_{\mathrm{RER}}
=
\frac{
A_{\mathrm{OH}}^{\mathrm{RL}} - A_{\mathrm{OH}}^{\mathrm{Base}}
}{
1 - A_{\mathrm{OH}}^{\mathrm{Base}}
}
-
\frac{
A_{\mathrm{EH}}^{\mathrm{RL}} - A_{\mathrm{EH}}^{\mathrm{Base}}
}{
1 - A_{\mathrm{EH}}^{\mathrm{Base}}
}.
\label{eq:s_rer}
\end{equation}
Here, $A_g^{\mathrm{RL}}$ and $A_g^{\mathrm{Base}}$ denote the mean accuracies of the RL-trained and corresponding base models on category $g$, respectively.
Lower $S_{\mathrm{RER}}$ indicates more EH-oriented improvement relative to OH.

\paragraph{External benchmarks.}
We evaluate visual hallucination using HallusionBench~\cite{guan2024hallusionbench} and CHAIR~\cite{rohrbach2018object}, and general multimodal capabilities using MMMU~\cite{yue2024mmmu} and OCRBench v2~\cite{fu2026ocrbench}.
We report strict binary accuracy on HallusionBench, sentence-level CHAIR$_s$ on COCO validation images, and accuracy on the MMMU.
For OCRBench v2, we report the macro-averaged score over English task categories.

\paragraph{Models and baselines.}
We evaluate \textbf{Qwen2.5-VL-3B}~\cite{Qwen2VL,qwen2.5-VL}, \textbf{Qwen2.5-VL-7B}~\cite{Qwen2VL,qwen2.5-VL}, and \textbf{LLaVA-OneVision-1.5-8B}~\cite{an2025llavaonevision15fullyopenframework,xie2025region} on eight NVIDIA RTX 5880 GPUs.
For each model, we compare the base model, DAPO~\cite{yu2026dapo}, fixed-weight DAPO, and our risk-constrained method.
Additional experiments on LLaVA-OneVision-1.5-8B transfer the verifiability-aware reward to PPO~\cite{schulman2017proximal} and GRPO~\cite{shao2024deepseekmath} to examine its applicability beyond DAPO.

\subsection{Main Results}
\label{sec:main_results}
\begin{table*}[!htb]
  \centering
  \caption{\textbf{Performance comparisons between DAPO-V and baselines on ITHV test set.}
  For fixed-weight DAPO, we report two representative settings.
  The boldface represents the best performance, and the underline represents the second-best.
  A dash indicates that $S_{\mathrm{RER}}$ is not applicable to the base model, which serves as the reference.}
  \label{tab:main-results-table}
  \small
  \setlength{\tabcolsep}{4.5pt}
  \renewcommand{\arraystretch}{1.18}
  \begin{tabularx}{\textwidth}{
  >{\raggedright\arraybackslash}p{0.17\textwidth}
  >{\raggedright\arraybackslash}p{0.31\textwidth}
  *{5}{>{\centering\arraybackslash}X}
}
    \toprule
    \multicolumn{1}{c}{\multirow{2}{*}{\textbf{Model}}}
      & \multicolumn{1}{c}{\multirow{2}{*}{\textbf{Method}}}
      & \multicolumn{5}{c}{\textbf{Hallucination Verifiability}} \\
    \cmidrule(lr){3-7}
      & & \mbox{Overall $\uparrow$}
      & \mbox{OH $\uparrow$}
      & \mbox{EH $\uparrow$}
      & \mbox{$L_{\mathrm{H}}\downarrow$}
      & \mbox{$S_{\mathrm{RER}}\downarrow$} \\
    \midrule
    \cellcolor{white} & Base & 57.51 & 50.45 & 27.94 & 23.60 & --- \\
    \rowcolor{gray!7}
    \cellcolor{white} & $+$ DAPO & 69.96 & 76.58 & 44.49 & 16.90 & 29.77 \\
    \cellcolor{white} & $+$ Fixed-weight DAPO ($\lambda=0.5$) & 72.25 & \textbf{78.15} & 47.06 & 15.84 & 29.37 \\
    \rowcolor{gray!7}
    \cellcolor{white} & $+$ Fixed-weight DAPO ($\lambda=1$) & \textbf{73.63} & \underline{77.93} & \underline{50.74} & \underline{15.34} & \underline{23.82} \\
    \rowcolor{blue!7}
    \cellcolor{white}\multirow{-5}{=}{\raggedright\textbf{Qwen2.5-VL-3B}} & \textbf{$+$ DAPO-V (Ours)} & \underline{72.75} & 77.70 & \textbf{54.21} & \textbf{14.72} & \textbf{18.54} \\
    \midrule
    \cellcolor{white} & Base & 66.85 & 65.99 & 37.87 & 19.39 & --- \\
    \rowcolor{gray!7}
    \cellcolor{white} & $+$ DAPO & 77.82 & 82.56 & 58.93 & 12.40 & 14.82 \\
    \cellcolor{white} & $+$ Fixed-weight DAPO ($\lambda=0.5$) & \textbf{78.75} & \textbf{83.33} & \underline{60.66} & \underline{12.09} & 14.30 \\
    \rowcolor{gray!7}
    \cellcolor{white} & $+$ Fixed-weight DAPO ($\lambda=1$) & 77.47 & 82.66 & 59.56 & 12.43 & \underline{14.10} \\
    \rowcolor{blue!7}
    \cellcolor{white}\multirow{-5}{=}{\raggedright\textbf{Qwen2.5-VL-7B}} & \textbf{$+$ DAPO-V (Ours)} & \underline{78.30} & \underline{82.88} & \textbf{64.71} & \textbf{11.15} & \textbf{6.46} \\
    \midrule
    \cellcolor{white} & Base & 60.71 & 53.38 & 32.72 & 23.21 & --- \\
    \rowcolor{gray!7}
    \cellcolor{white} & $+$ DAPO & 77.34 & 80.63 & 56.82 & 14.60 & 22.63 \\
    \cellcolor{white} & $+$ Fixed-weight DAPO ($\lambda=0.5$) & 77.38 & \underline{80.86} & 56.99 & 13.97 & 22.87 \\
    \rowcolor{gray!7}
    \cellcolor{white} & $+$ Fixed-weight DAPO ($\lambda=1$) & \underline{78.53} & 80.63 & \underline{63.81} & \underline{12.26} & \textbf{12.24} \\
    \rowcolor{blue!7}
    \cellcolor{white}\multirow{-5}{=}{\raggedright\textbf{\shortstack[l]{LLaVA-OneVision-\\1.5-8B}}} & \textbf{$+$ DAPO-V (Ours)} & \textbf{79.04} & \textbf{81.98} & \textbf{65.13} & \textbf{11.75} & \underline{13.18} \\
    \bottomrule
  \end{tabularx}
  \par\vspace{4pt}
\end{table*}

Figure~\ref{fig:obvious_vs_elusive} and Table~\ref{tab:main-results-table} summarize hallucination mitigation results across three MLLMs.
Although DAPO~\cite{yu2026dapo} improves accuracy on both OH and EH, substantial room remains for EH mitigation.
Fixed verifiability-based weighting improves EH performance in several settings, but increasing its coefficient does not consistently yield larger accuracy gains.
Among all compared methods, DAPO-V achieves the highest EH gains and lowest verifiability-aware risk $L_H$ for every base model.
Relative to DAPO, it improves EH gains by 5.78--9.72 percentage points and reduces $L_H$ by 10.1--19.5\%, while also improving overall accuracy.
The corresponding decrease in $S_{\mathrm{RER}}$ across all three models indicates a shift toward greater relative improvement on EH.
Compared with fixed weighting, these gains reflect a trade-off rather than uniform dominance: fixed-weight baselines attain higher overall accuracy on the two Qwen models, whereas DAPO-V achieves higher EH accuracy and lower $L_H$.
On LLaVA-OneVision-1.5-8B, DAPO-V also achieves the highest overall accuracy.
These results support the effectiveness of risk-constrained learning for mitigating errors on instances estimated to be harder for humans to verify.

\subsection{Analysis}

\begin{figure*}[!htb]
    \centering
    \includegraphics[width=0.97\textwidth]{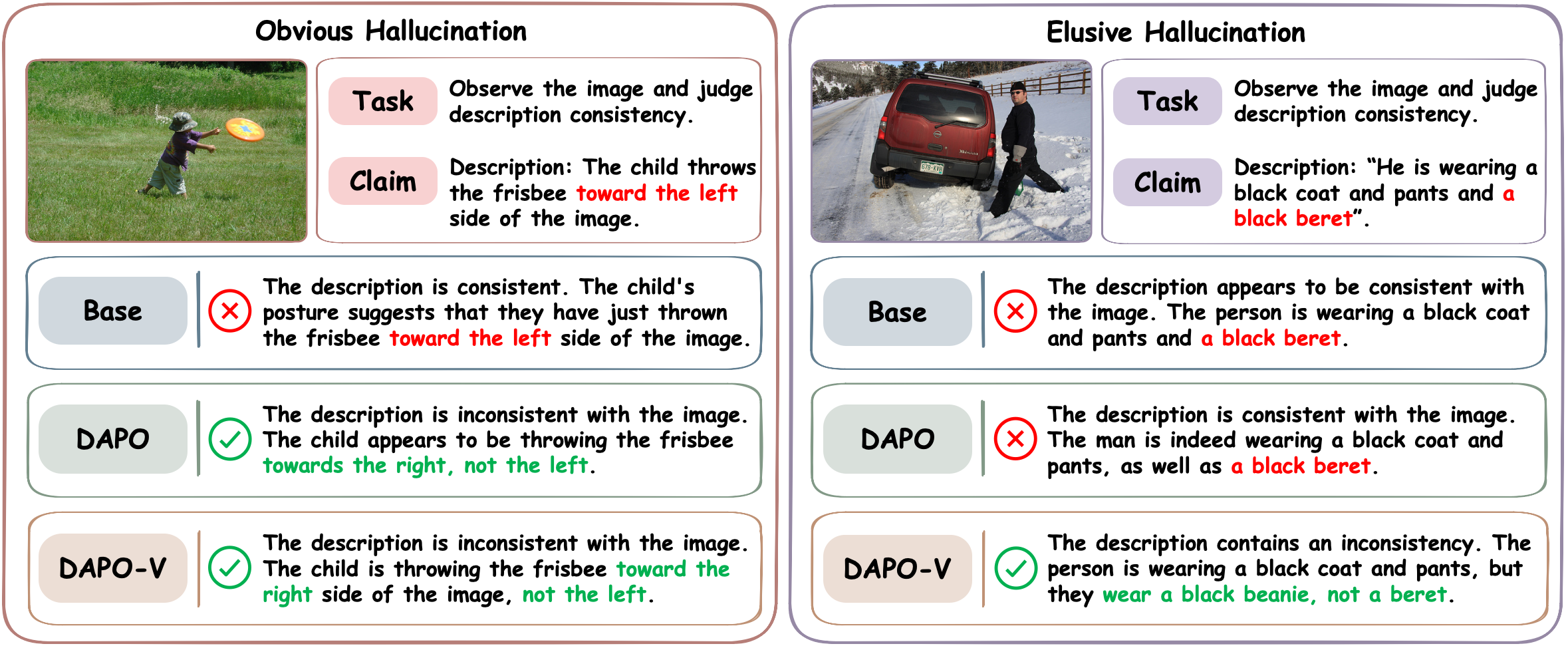}
    \caption{\textbf{Qualitative comparison of obvious and elusive hallucination mitigation on LLaVA-OneVision-1.5-8B.}
    Hallucinated content is highlighted in red, and corrected or image-grounded content is highlighted in green.}
    \label{fig:case_study}
\end{figure*}

\paragraph{Effects of the risk budget.}
Figure~\ref{fig:risk_budget} evaluates different risk budgets on the validation set.
Across three models, $\delta=0.1$ yields the highest EH accuracy and lowest $L_H$.
The corresponding EH accuracies are 55.88\%, 61.03\%, and 63.76\% for Qwen2.5-VL-3B, Qwen2.5-VL-7B, and LLaVA-OneVision-1.5-8B, respectively.
Qwen2.5-VL-3B also achieves its lowest $S_{\mathrm{RER}}$ at this budget, whereas the two larger models minimize $S_{\mathrm{RER}}$ at $\delta=0.2$.
Compared with zero-budget control, $\delta=0.1$ improves all three metrics across all models, supporting a moderate positive budget for balancing EH accuracy and verifiability-aware risk.

\paragraph{Case study.}
Figure~\ref{fig:case_study} compares Base, DAPO, and DAPO-V on OH and EH examples.
For OH, Base accepts the incorrect leftward-throw claim, whereas DAPO and DAPO-V correctly identify the rightward direction.
For EH, Base and DAPO repeat the incorrect ``black beret'' claim, while DAPO-V identifies a black beanie and preserves the correct coat and pants descriptions.
Thus, DAPO corrects the salient directional error but misses the subtle hat-type error, whereas DAPO-V resolves both.
This illustrates how our objective penalizes harder-to-verify errors more heavily while preserving correction of obvious ones.

\noindent
\begin{minipage}[t]{0.47\linewidth}
    \vspace{0pt}
    \paragraph{Human evaluation.}
    To assess the detectability of model-generated errors, we conducted a blinded human evaluation.
    We used LLaVA-OneVision-1.5-8B to generate image descriptions under the Base, DAPO, and DAPO-V settings, using the same images and prompt (shown in Appendix~\ref{app:human_study_prompt}).
    We screened the descriptions against the images to identify erroneous content.
    Ten participants then evaluated 100 hallucinated image--sentence pairs per model, judging each pair as consistent or inconsistent without knowing its model source.
\end{minipage}\hfill
\begin{minipage}[t]{0.49\linewidth}
    \vspace{0pt}
    \centering
\captionsetup{type=table}
\caption{Blinded human evaluation of hallucinations generated by LLaVA-OneVision-1.5-8B. DAPO-V yields the highest EH accuracy and detection rate, as well as the shortest response time.}
\label{tab:human_evaluation}
\small
\begin{tabularx}{\linewidth}{l >{\centering\arraybackslash}X c >{\centering\arraybackslash}X}
\toprule
Model & EH & Detection (\%) & Time (s) \\
\midrule
\rowcolor{white}
Base   & 32.72 & 70.00 & 12.39 \\

\rowcolor{gray!7}
DAPO   & 56.82 & 60.00 & 13.15 \\

\rowcolor{blue!7}
\textbf{DAPO-V} & \textbf{65.13} & \textbf{77.20} & \textbf{11.84} \\
\bottomrule
\end{tabularx}

\end{minipage}

Both detection rates and mean response times are computed on hallucinated image--sentence pairs.
As shown in Table~\ref{tab:human_evaluation}, DAPO-V achieves a hallucination detection rate of 77.2\%, exceeding Base and DAPO by 7.2\% and 17.2\%.
It also yields the shortest mean response time (11.84s, versus 12.39s for Base and 13.15s for DAPO).
These results suggest that DAPO-V generates hallucinations that are easier for humans to detect.

\noindent
\begin{minipage}[t]{0.48\linewidth}
    \vspace{0pt}
    {
        \centering
        \captionsetup{type=figure}
        \includegraphics[width=\linewidth]{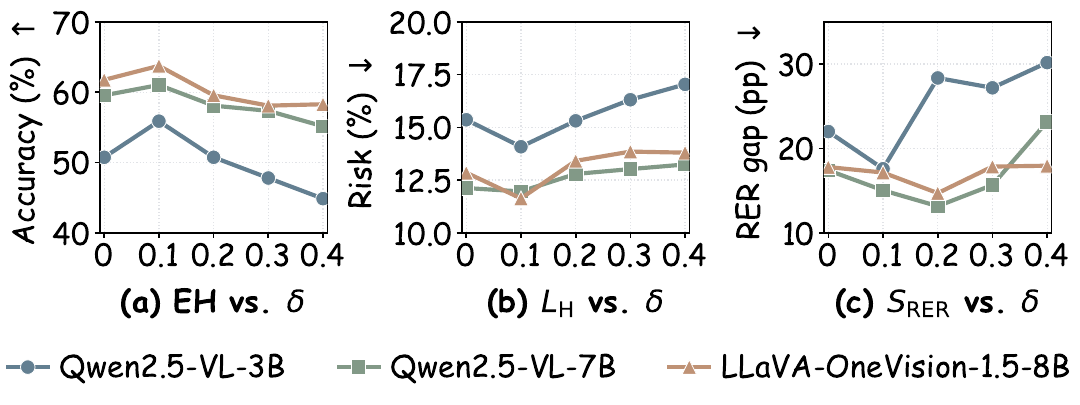}
        \caption{\textbf{Sensitivity to the risk budget $\delta$.}
        (a) EH accuracy, (b) verifiability-aware risk $L_H$, and (c) relative error-reduction gap $S_{\mathrm {RER}}$ across the three models as $\delta$ varies.}
        \label{fig:risk_budget}
        \par
    }

    \paragraph{Hallucination mitigation and general capabilities.}
    To assess whether gains on hallucination-prone questions come at the expense of broader model capabilities, we evaluate visual hallucination and general utility on external benchmarks (Table~\ref{tab:benchmark-comparison}).
\end{minipage}\hfill
\begin{minipage}[t]{0.48\linewidth}
    \vspace{0pt}
    
  \centering
  \captionsetup{type=table}
  \caption{Performance comparisons between DAPO-V and baselines on multimodal benchmarks.}
  \label{tab:benchmark-comparison}
  \fontsize{7.5}{9}\selectfont
  \setlength{\tabcolsep}{2pt}
  \renewcommand{\arraystretch}{1.08}

  \begin{tabularx}{\linewidth}{
    >{\raggedright\arraybackslash}X
    cccc
  }
    \toprule
    \multicolumn{1}{>{\centering\arraybackslash}X}
    {\multirow{2}{*}{\textbf{Method}}}
      & \multicolumn{2}{c}{\textbf{Visual Hallucination}}
      & \multicolumn{2}{c}{\textbf{General Utility}} \\
    \cmidrule(lr){2-3}
    \cmidrule(lr){4-5}
      & HallusionBench~$\uparrow$
      & CHAIR~$\downarrow$
      & MMMU~$\uparrow$
      & OCRBench v2~$\uparrow$ \\
    \midrule

    \multicolumn{5}{l}{\textbf{Qwen2.5-VL-3B}} \\
    \midrule

    Base
      & 64.22 & 36.80 & 50.00 & 46.22 \\

    \rowcolor{gray!7}
    DAPO
      & 67.67 & 34.20 & 49.67 & 46.02 \\

    \rowcolor{blue!7}
    \textbf{DAPO-V}
      & \textbf{68.82}
      & \textbf{33.80}
      & \textbf{51.11}
      & \textbf{46.62} \\

    \midrule

    \multicolumn{5}{l}{\textbf{Qwen2.5-VL-7B}} \\
    \midrule

    Base
      & 70.50 & 33.40 & 52.78 & 46.00 \\

    \rowcolor{gray!7}
    DAPO
      & 69.80 & \textbf{30.40} & 53.67 & 45.61 \\

    \rowcolor{blue!7}
    \textbf{DAPO-V}
      & \textbf{71.48}
      & 31.00
      & \textbf{53.89}
      & \textbf{46.18} \\

    \midrule

    \multicolumn{5}{l}{\textbf{LLaVA-OneVision-1.5-8B}} \\
    \midrule

    Base
      & 65.01 & 29.60 & 55.89 & 48.15 \\

    \rowcolor{gray!7}
    DAPO
      & 69.26 & 28.80 & 56.44 & 48.39 \\

    \rowcolor{blue!7}
    \textbf{DAPO-V}
      & \textbf{69.53}
      & \textbf{28.20}
      & \textbf{56.56}
      & \textbf{48.40} \\

    \bottomrule

  \end{tabularx}

\end{minipage}

DAPO-V achieves higher HallusionBench scores across three models and lower CHAIR scores on two of them, demonstrating improved hallucination mitigation beyond ITHV.
It also consistently outperforms DAPO on general utility across all three models.

\noindent
\begin{minipage}[t]{0.53\linewidth}
    \vspace{0pt}
    \paragraph{Generalization to other RL algorithms.}
    We further ask whether the verifiability-aware reward benefits policy optimizers beyond DAPO.
    We transfer the reward and its adaptive penalty to PPO and GRPO, obtaining \textbf{PPO-V} and \textbf{GRPO-V}.
    Neither uses DAPO's specialized pre-filtering risk monitor: without dynamic group filtering, risk is estimated directly from the unfiltered rollout batch.
    As shown in Table~\ref{tab:rl_comparison}, PPO-V and GRPO-V improve EH accuracy by 4.10 and 5.18 percentage points and reduce $L_H$ by 8.6\% and 12.2\%, respectively, with lower $S_{\mathrm{RER}}$ in both comparisons.
    These results support the transferability of the reward mechanism, while pre-filtering monitoring remains a DAPO-specific integration component.
\end{minipage}\hfill
\begin{minipage}[t]{0.43\linewidth}
    \vspace{0pt}
    
  \centering
  \captionsetup{type=table}
  \caption{Transfer of the verifiability-aware reward to PPO and GRPO on LLaVA-OneVision-1.5-8B.}
  \label{tab:rl_comparison}
  \small
  \setlength{\tabcolsep}{8pt}
  \begin{tabular}{lccc}
    \toprule
    \multicolumn{1}{c}{Method} & EH $\uparrow$ & $L_H$ $\downarrow$ & $S_{\mathrm{RER}}$ $\downarrow$ \\
    \midrule

    \rowcolor{white}
    PPO   & 51.94 & 15.65 & 10.09 \\

    \rowcolor{blue!7}
    \textbf{PPO-V} & \textbf{56.04} & \textbf{14.31} & \textbf{9.14} \\
    \midrule

    \rowcolor{white}
    GRPO   & 51.57 & 15.78 & 20.22 \\

    \rowcolor{blue!7}
    \textbf{GRPO-V} & \textbf{56.75} & \textbf{13.86} & \textbf{15.76} \\
    \midrule

    \rowcolor{white}
    DAPO   & 56.82 & 14.60 & 22.63 \\

    \rowcolor{blue!7}
    \textbf{DAPO-V} & \textbf{65.13} & \textbf{11.75} & \textbf{13.18} \\
    \bottomrule
  \end{tabular}

\end{minipage}

\subsection{Ablation Study}

Table~\ref{tab:ablation_results} evaluates the contributions of verifiability-aware penalties, the risk-budget offset, and pre-filtering risk monitoring across the three models.

\noindent
\begin{minipage}[t]{0.49\linewidth}
    \vspace{0pt}
    \centering
\captionsetup{type=table}
\caption{
Ablation study of DAPO-V across three models, examining the effects of verifiability-aware penalties, the risk-budget offset, and pre-filtering risk monitoring.
}
\label{tab:ablation_results}
\footnotesize
\setlength{\tabcolsep}{2.5pt}
\renewcommand{\arraystretch}{1.15}

\begin{tabularx}{\linewidth}{
    >{\raggedright\arraybackslash}X
    ccc
}
    \toprule
    \multicolumn{1}{>{\centering\arraybackslash}X}
    {\textbf{Method}}
    & \textbf{EH $\uparrow$}
    & \textbf{$L_H$ $\downarrow$}
    & \textbf{$S_{\mathrm{RER}}$ $\downarrow$} \\
    \midrule

    \multicolumn{4}{l}{\textbf{Qwen2.5-VL-3B}} \\
    \midrule

    DAPO-V w/o verifiability-aware penalty
    & 42.28 & 17.35 & 36.00 \\

    \rowcolor{gray!7}
    DAPO-V w/o budget offset
    & 51.10 & 15.22 & 25.14 \\

    DAPO-V w/o Pre-filtering monitoring
    & 52.21 & 15.41 & 27.23 \\

    \rowcolor{blue!7}
    \textbf{DAPO-V}
    & \textbf{54.21}
    & \textbf{14.72}
    & \textbf{18.54} \\

    \midrule

    \multicolumn{4}{l}{\textbf{Qwen2.5-VL-7B}} \\
    \midrule

    DAPO-V w/o verifiability-aware penalty
    & 57.35 & 12.96 & 17.66 \\

    \rowcolor{gray!7}
    DAPO-V w/o budget offset
    & 59.93 & 12.04 & 15.48 \\

    DAPO-V w/o Pre-filtering monitoring
    & 55.88 & 13.13 & 18.03 \\

    \rowcolor{blue!7}
    \textbf{DAPO-V}
    & \textbf{64.71}
    & \textbf{11.15}
    & \textbf{6.46} \\

    \midrule

    \multicolumn{4}{l}{\textbf{LLaVA-OneVision-1.5-8B}} \\
    \midrule

    DAPO-V w/o verifiability-aware penalty
    & 55.88 & 14.27 & 24.03 \\

    \rowcolor{gray!7}
    DAPO-V w/o budget offset
    & 60.03 & 13.07 & 15.65 \\

    DAPO-V w/o Pre-filtering monitoring
    & 59.40 & 13.38 & 16.87 \\

    \rowcolor{blue!7}
    \textbf{DAPO-V}
    & \textbf{65.13}
    & \textbf{11.75}
    & \textbf{13.18} \\

    \bottomrule
\end{tabularx}

\end{minipage}\hfill
\begin{minipage}[t]{0.47\linewidth}
    \vspace{0pt}
    \paragraph{Verifiability-aware penalty.}
    This variant removes the verifiability-aware penalty but retains DAPO-V's mean-centered advantages without standard-deviation normalization, distinguishing it from standard DAPO in Table~\ref{tab:main-results-table}.
    Removing the verifiability-aware penalty reduces EH accuracy by 11.93, 7.36, and 9.25 percentage points on Qwen2.5-VL-3B, Qwen2.5-VL-7B, and LLaVA-OneVision-1.5-8B.
    It also increases $L_H$ and $S_{\mathrm{RER}}$ for every model.
    These results support incorporating human verifiability into policy optimization rather than optimizing correctness alone.

    \smallskip
    \paragraph{Risk-budget offset.}
    Setting $\delta=0$ removes the budget offset while retaining adaptive multiplier updates, yielding zero-budget control.
    Compared with DAPO-V, this variant decreases EH accuracy by 3.11--5.10 percentage points and increases $L_H$ across all models.
    The nonzero budget provides a more effective operating point in these experiments.
\end{minipage}

\paragraph{Pre-filtering risk monitoring.}
We replace pre-filtering monitoring with risk estimates computed only from DAPO-retained groups within the same raw monitoring window, leaving the policy-training selection rule unchanged.
This replacement decreases EH accuracy by 2.00--8.83 percentage points and increases $L_H$ for every model.
These findings support separating risk monitoring from correctness-dependent training-group selection.

\Needspace{6\baselineskip}
\section{Related Work}
\label{sec:related_work}

\subsection{Hallucination of Multimodal Language Models}

Multimodal language models can generate linguistically plausible statements unsupported by visual evidence~\cite{guo2026seeing}.
Evaluation protocols quantify object hallucination~\cite{rohrbach2018object,li2023evaluating} and extend to attribute and relation grounding or image-context reasoning~\cite{wang2023amber,guan2024hallusionbench}.
Mitigation methods intervene during decoding~\cite{leng2024mitigating,huang2024opera}, revise hallucinated descriptions after generation~\cite{zhou2024analyzing}, or improve instruction tuning with positive and negative examples~\cite{liu2024mitigating}.
Feedback-based alignment exploits factual reward modeling, fine-grained corrections, and AI-generated supervision~\cite{sun2024aligning,yu2024rlhf,yu2025rlaif}, while preference-optimization methods construct hallucination-focused training pairs or preserve image conditioning~\cite{zhou2024aligning,wang2024mdpo}, while OPA-DPO addresses alignment between preference data and the reference policy~\cite{yang2025mitigating}.
Recent work investigates hallucination in multimodal RL post-training~\cite{Zhang_2026_CVPR}.
Constrained multimodal alignment has also been explored for safety~\cite{ji2026safe}.
In contrast, our work constrains factual errors according to human verification difficulty.

\subsection{Verifiability of AI-generated Content}

Fluent model outputs can appear authoritative despite factual errors~\cite{zhou2024relying,kim2025fostering}.
Studies show that convincing explanations can promote acceptance of incorrect answers and inflate confidence without improving accuracy~\cite{si2024large,steyvers2025large,ajwani2024llm}.
Linguistic plausibility is therefore not a reliable substitute for verification;
it may instead increase the cognitive burden of human verification ~\cite{ji2023survey}.
Existing approaches include evidence-linked generation~\cite{nakano2021webgpt,menick2022teaching}, attribution evaluation~\cite{gao2023enabling,liu2023evaluating}, calibrated uncertainty~\cite{lin2022teaching}, and checkability-oriented training~\cite{anil2021learning,kirchner2024prover}.
In multimodal settings, prior work distinguishes obvious from elusive hallucinations using human verification outcomes and regulates them through activation steering~\cite{pang2026steering}.
We instead constrain verifiability-aware risk during policy learning.
\section{Conclusion}
\label{sec:conclusion}

Improved RL task performance does not necessarily make residual hallucinations easier for humans to verify.
To study hallucinations across levels of human verification difficulty, we construct ITHV, a dataset of 2,728 visually grounded instances that uses human judgments to distinguish obvious from elusive hallucinations.
Building on ITHV, we propose DAPO-V to optimize correctness subject to a verifiability-aware risk budget.
DAPO-V penalizes incorrect responses more heavily on questions with lower human verification accuracy and uses adaptive dual control to adjust the penalty strength in response to budget violations.
Across three MLLMs, DAPO-V improves accuracy both overall and on elusive-hallucination questions compared with accuracy-only DAPO.
A blinded human study provides evidence of improved human verifiability of the generated hallucinations from DAPO-V.
PPO and GRPO experiments support the transferability of the verifiability-aware reward beyond DAPO.
Our findings motivate incorporating human verifiability into RL post-training alongside correctness.

\clearpage

\bibliographystyle{plainnat}
\bibliography{main}

\clearpage

\newpage
\beginappendix


\section{Additional Details of ITHV Dataset}
\label{app:dataset_details}
\paragraph{Image sources and task construction.}
\label{app:ithv_sources}
We draw images from AMBER~\cite{wang2023amber} and RealWorldQA~\cite{realworldqa2024}.
Each instance contains an image, a question, and a reference answer.
Verification tasks require identifying visual inconsistencies and, where requested, correcting unsupported claims; question-answering tasks require image-grounded answers.

\subsection{Participants and Annotation Protocol}
\label{app:dataset_participants}
We recruited 65 volunteers through social media, aged 18–60, with diverse academic backgrounds including Mathematics, Design, and Computer Science. 
This diversity was intended to incorporate perspectives from different academic disciplines rather than relying on a single field. 
This study was approved by the ethics committee (anonymized during revision).
Each source item was independently assessed by five participants under a 20-second time limit.
Participants were asked to evaluate the visual consistency of each image–text pair.
Upon reaching the 20-second limit, the system automatically submitted the recorded answer.
Automatically submitted answers were marked as correct if the selected answer was correct; otherwise, incorrect or unanswered responses were counted as incorrect.
Invalid responses were excluded before aggregation.
The resulting human accuracy and response-time measurements were used to characterize item-level verification difficulty and support the OH/EH categorization described in Section~\ref{sec:ITHV_annotation}.

\subsection{Sensitivity to the Response-Time Threshold}
We assess the sensitivity of category assignments to the median response-time threshold $\tau \in \{8,10,12,14,16\}$ seconds.
The analysis uses a fixed pool of 2,036 candidates, comprising 1,829 OH/EH instances under the default 12-second threshold and 207 additional instances with $a_i=0.6$ and median response times not exceeding 12 seconds.
For each threshold, we retain the original accuracy-based criteria and reassign the borderline instances according to whether their median response time exceeds $\tau$.
Relative to the default 12-second threshold, the proportions of candidates whose labels change are 4.86\%, 2.01\%, 0.93\%, and 2.46\% at 8, 10, 14, and 16 seconds, respectively.
These results indicate that the OH/EH categorization is relatively robust to the response-time threshold, with label changes affecting at most 4.86\% of the fixed candidate pool across the evaluated settings.

\subsection{Detailed Dataset Statistics}
\label{app:dataset_analysis}

\begin{figure*}[t]
    \centering
    \includegraphics[width=\textwidth]{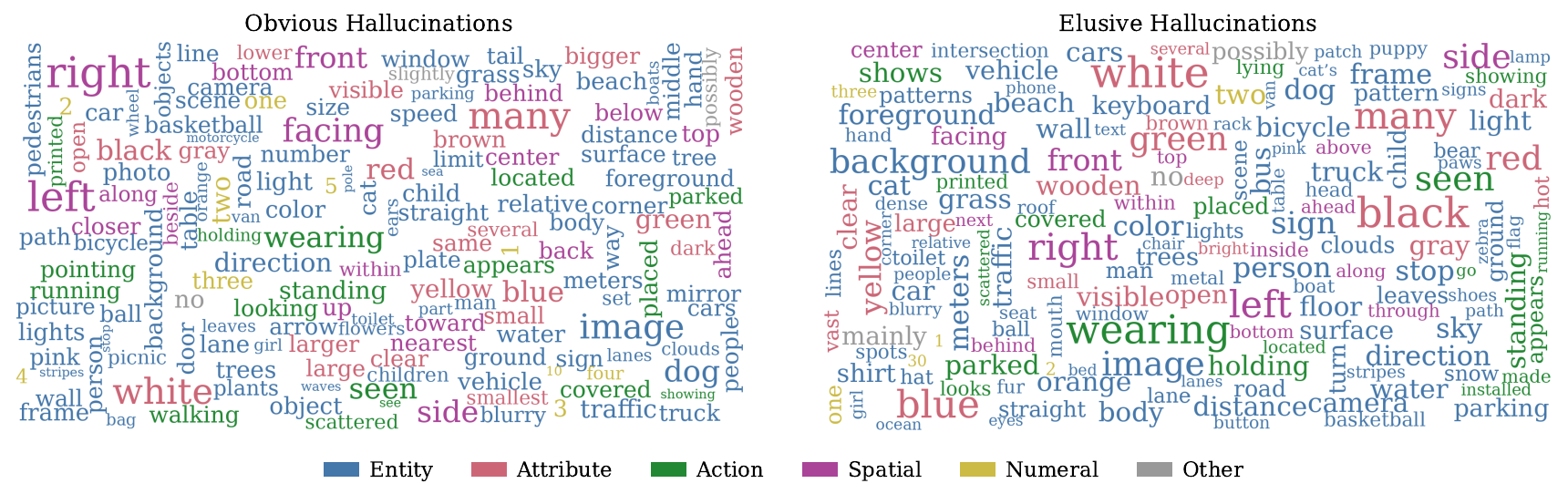}
    \caption{
    English-normalized verification-target vocabulary for
    (a) obvious and (b) elusive hallucinations. Font size is
    proportional to target frequency, and color indicates the
    corresponding linguistic--semantic category.
    }
    \label{fig:english_wordcloud}
\end{figure*}

To complement the aggregate category distributions in Figure~\ref{fig:dataset}(c)(ii), Figure~\ref{fig:english_wordcloud} provides a qualitative visualization of the verification-target vocabulary. 
Each verification target denotes the minimal word or phrase whose visual grounding is central to judging the corresponding sample.
For visualization, the targets are mapped to English-normalized surface forms. Font size indicates frequency. 
Both subsets cover diverse entities, attributes, actions, spatial relations, and numerical or textual targets, with substantial vocabulary overlap. 
This suggests that the verifiability labels are not determined by expression type, but are derived from aggregated human judgments.

We construct image-disjoint training, validation, and test splits by grouping all instances associated with the same source image and assigning each group to a single split.
All questions derived from an image are kept together, ensuring that no source image is shared across splits.
We use seed 42 for this splitting procedure.

\subsection{Qualitative Analysis of Human Verifiability}
\label{app:dataset_example}
Figure~\ref{fig:dataset}(b) illustrates how claims about the same image can differ in human verifiability.
For the boy-and-goose image, participants reject the nonexistent toy gun with 100\% accuracy, whereas the claim misidentifying the pacifier as a blue candy is rejected with only 40\% accuracy.
Both involve entity verification, but different human verification outcomes.
Similarly, for the boat image, the cloudless-sky claim is identified as incorrect with 80\% accuracy, compared with 20\% for the claim suggesting a wooden boat.
These cases suggest that plausible object substitutions, material inferences and fine-grained recognition can be less readily detected than direct visual contradictions.
They illustrate differences in verifiability that broad semantic categories alone do not capture.
Together with the within-category variation in human accuracy, these observations motivate retaining sample-level accuracy as a proxy for human verifiability rather than relying solely on OH/EH labels.

\section{Verification and Optimization Details}
\label{app:method_details}

\subsection{Correctness Verification}
\label{app:verification}

The verifier first compares normalized complete responses with the human-reviewed reference.
Normalization applies Unicode normalization, whitespace normalization, case folding, and limited sentence-final punctuation handling; it does not extract an answer label while discarding additional claims.
Exact matches are accepted directly.
Other responses are evaluated by Qwen3.5-9B~\cite{qwen3.5}, which serves as a frozen judge conditioned on the image, question, reference answer, and complete candidate response.
Semantically equivalent answers are accepted, whereas false additional claims, contradictions, unnecessary refusals, and incomplete task fulfillment are rejected.
For correction tasks, both the consistency decision and the requested factual repair must be correct.
The judge returns a single Boolean correctness field, without a rationale-based reward.

\subsection{Policy Optimization Details}
\label{app:policy_optimization}

To match the advantage scale of the fixed-weight baseline, we rescale the relative advantages in Eq.~\eqref{eq:remove_norm}:
\begin{equation}
A^{\mathrm{opt}}_{ij}=2\widehat A_{ij}.
\label{eq:app_advantage_scale}
\end{equation}
The factor of two is determined by the baseline's symmetric
reward encoding rather than tuned.
Specifically, under mean centering without standard-deviation
normalization, matched coefficients
$\lambda_{\mathrm{fix}}=\lambda_t/p_H$ yield
\begin{equation}
\widehat A^{\mathrm{fix}}_{ij}
=2u_i(C_{ij}-\bar C_i)
=2\left(1+\frac{\lambda_t}{p_H}w_i\right)
(C_{ij}-\bar C_i)
=A^{\mathrm{opt}}_{ij}.
\end{equation}
This fixed rescaling preserves relative verifiability-based
weighting and leaves the dual update unchanged.

For token $k$ of response $y_{ij}$, define the probability ratio
\begin{equation}
\rho_{ijk}(\theta)
=
\frac{\pi_\theta(y_{ij,k}\mid x_i,y_{ij,<k})}
{\pi_{\theta_t}(y_{ij,k}\mid x_i,y_{ij,<k})},
\end{equation}
where $\pi_{\theta_t}$ is the policy used to collect the current rollouts.
Following DAPO~\cite{yu2026dapo}, the asymmetric clipped surrogate is
\begin{equation}
b(\rho,A)=\min\bigl\{
\rho A,\,
\operatorname{clip}(\rho,1-\epsilon_{\mathrm{low}},1+\epsilon_{\mathrm{high}})A
\bigr\}.
\end{equation}
The implementation additionally retains a negative-advantage dual-clipping safeguard:
\begin{equation}
s(\rho,A)=
\begin{cases}
\max\{b(\rho,A),\kappa A\},&A<0,\\
b(\rho,A),&A\geq0.
\end{cases}
\end{equation}
Here, $\kappa$ is a clipping constant, unrelated to the multiplier $\lambda_t$.
For a retained response batch $\mathcal B_t$, we minimize
\begin{equation}
\mathcal L_{\mathrm{policy}}(\theta)
=
-\frac{
\displaystyle\sum_{(i,j)\in\mathcal B_t}
\sum_{k=1}^{|y_{ij}|}s\bigl(\rho_{ijk}(\theta),A^{\mathrm{opt}}_{ij}\bigr)
}{
\displaystyle\sum_{(i,j)\in\mathcal B_t}|y_{ij}|
}.
\label{eq:app_policy_loss}
\end{equation}
Sequence lengths count response tokens only, and the
denominator includes all retained responses.
We use $\epsilon_{\mathrm{low}}=0.2$, $\epsilon_{\mathrm{high}}=0.28$, and $\kappa=10$, without a reference-policy KL penalty.

\paragraph{Control and batch settings.}
Unless otherwise specified, $G=8$, $M=66$, $\beta=0.8$, $\eta_\lambda=0.1$, and $\lambda_{\max}=1$.
The dual multiplier is initialized as $\lambda_0=0.5p_H$ for all DAPO-V experiments, unless otherwise specified..
We retain groups with $0.01<\bar C_i<0.99$.
Each outer iteration collects 63 retained question groups and performs three optimizer updates, each using 21 groups of eight responses.
The multiplier is fixed for all these updates and adjusted once at the outer-iteration boundary.
All RL variants update language-side LoRA adapters with rank 64 and scaling parameter 64.
The learning rate is $10^{-5}$ with a warmup ratio of 0.03.

\subsection {Fixed-Weight Baselines}
\label {sec:fixed_weight_baselines}
We compare DAPO-V with fixed-weight DAPO, which incorporates human verification difficulty through a constant reward coefficient without risk-budget feedback.
Let $\lambda_{\mathrm{fix}}\geq 0$ denote this coefficient, distinguished from the adaptive multiplier $\lambda_t$ in DAPO-V. The instance weight is
\begin{equation}
u_i =
\begin{cases}
1+\lambda_{\mathrm{fix}}(1-a_i), & i\in\mathcal H,\\
1, & i\notin\mathcal H.
\end{cases}
\end{equation}
For response $y_{ij}$ with correctness $C_{ij}\in\{0,1\}$, the reward is
\begin{equation}
R_{ij}^{\mathrm{fix}}=u_i(2C_{ij}-1).
\end{equation}
Thus, correct and incorrect responses receive $+u_i$ and $-u_i$, respectively.
Instances without human accuracy receive unit weight rather than an imputed difficulty estimate.

For a group of $G$ responses, the mean-centered advantage is
\begin{equation}
\widehat A_{ij}^{\mathrm{fix}}
=
R_{ij}^{\mathrm{fix}}-\bar R_i^{\mathrm{fix}}
=
2u_i(C_{ij}-\bar C_i),
\end{equation}
where $\bar R_i^{\mathrm{fix}}=G^{-1}\sum_jR_{ij}^{\mathrm{fix}}$
and $\bar C_i=G^{-1}\sum_jC_{ij}$.
We use this advantage in policy optimization.

The baseline otherwise shares the same dataset, correctness verifier, LoRA configuration, and DAPO policy objective with DAPO-V.
Dynamic group selection uses raw correctness rather than weighted rewards.
We consider $\lambda_{\mathrm{fix}}\in\{0,0.25,0.5,0.75,1\}$. Unlike DAPO-V, these variants neither impose a risk budget nor update the coefficient during training.

\subsection{Additional RL Baselines and Reward Transfer}
\label{app:reward_transfer}

\paragraph{Shared protocol.}
We further evaluate reward transfer using LLaVA-OneVision-1.5-8B on the same ITHV splits.
Within each PPO/PPO-V or GRPO/GRPO-V pair, model initialization, data, and optimizer settings are matched.
Both families use language-side LoRA with rank and scaling parameter 64, actor learning rate $10^{-5}$, and warmup ratio 0.03, while keeping the pretrained base model and visual encoder frozen.

\paragraph{PPO and PPO-V.}
PPO~\cite{schulman2017proximal} uses a separately trained critic initialized from the same base model, with its own LoRA adapters and scalar value head.
The critic learning rate is $10^{-5}$.
Both variants use generalized advantage estimation with $\gamma=1$ and $\lambda_{\mathrm{GAE}}=0.95$, batch-level advantage normalization, and value clipping of 0.2.
The task reward is assigned to the final response token, with reference-policy KL regularization applied through token rewards.
PPO uses the correctness reward $C_{ij}$; PPO-V replaces it with the verifiability-aware reward defined in the main text.
The critic learns the resulting returns in both cases.

\paragraph{GRPO and GRPO-V.}
Both variants use critic-free group-relative updates~\cite{shao2024deepseekmath}.
The GRPO baseline uses $2C_{ij}-1$ and normalizes group-relative advantages by the within-group standard deviation.
GRPO-V instead uses the verifiability-aware reward and mean-centered advantages without standard-deviation normalization, preserving question-dependent weighting.
A fixed global scaling factor of two is applied to the GRPO-V advantages.
Reference-policy KL regularization is applied directly to the actor loss in both variants.
Neither variant performs DAPO-style dynamic filtering.

\paragraph{Adaptive risk control.}
PPO-V and GRPO-V transfer the verifiability-aware reward and adaptive multiplier, not the complete DAPO-V pipeline.
Risk is estimated from annotated prompts in the full, unfiltered rollout batch, without DAPO's specialized pre-filtering monitoring window.
The multiplier remains fixed within an outer iteration and is updated once after optimization.

Both families use one optimization pass per rollout batch, ratio clipping of 0.2.
The frozen reference policy is the corresponding pretrained base model with LoRA disabled.
Training configurations are matched within each baseline/\mbox{-V} pair.

\subsection{Training Dynamics}
\label{sec:training_dynamics}

\begin{figure}[!htb]
    \centering
    \begin{minipage}[t]{0.42\linewidth}
        \centering
        \includegraphics[width=\linewidth]{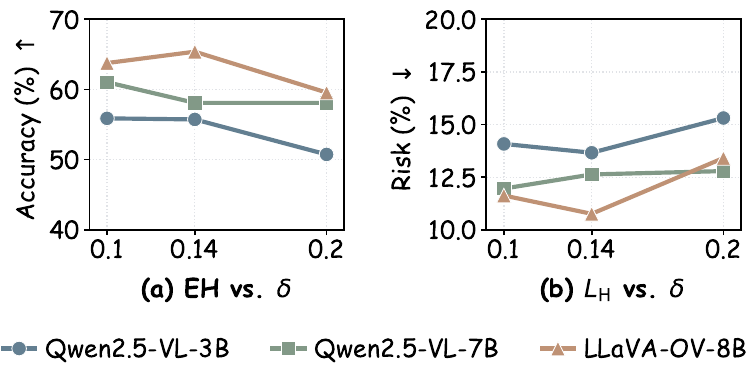}
        \caption{Local sensitivity to the risk budget on the validation set.
        EH accuracy and $L_H$ are evaluated at $\delta\in\{0.10,0.14,0.20\}$ using the same sampling protocol across budget settings.}
        \label{fig:local_risk_budget}
    \end{minipage}\hfill
    \begin{minipage}[t]{0.54\linewidth}
        \centering
        \includegraphics[width=\linewidth]{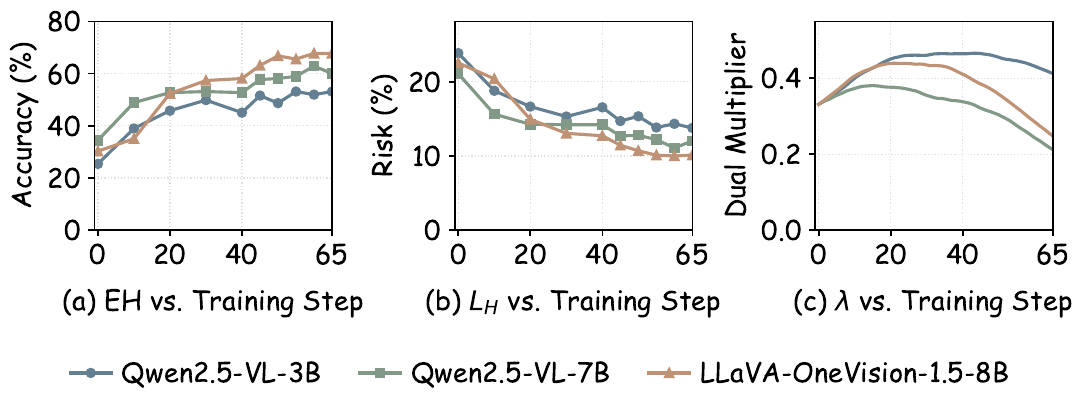}
        \caption{Training dynamics on ITHV with $\delta=0.14$.
        Panels (a) and (b) show EH accuracy and verifiability-aware risk $L_H$, measured on the ITHV validation set at training checkpoints.
        Panel (c) shows the dual multiplier $\lambda$ recorded during training.}
        \label{fig:training_dynamics}
    \end{minipage}
\end{figure}

For the three models, fixed-weight baselines with $\lambda_{\mathrm{fix}}\in\{0.5,1\}$ yield validation risks $L_H$ of approximately $0.13$--$0.15$.
The reference budget $\delta=0.14$ is therefore consistent with this empirical risk scale, rather than derived as an optimal threshold.
Figure~\ref{fig:local_risk_budget} further examines nearby budgets on the validation set, $\delta=0.14$ yields the lowest $L_H$ for Qwen2.5-VL-3B and the highest EH accuracy and lowest $L_H$ for LLaVA-OneVision-1.5-8B.
We use $\delta=0.14$ and $\lambda_0=0.5p_H$ as shared reference settings for the training-dynamics analysis.

\FloatBarrier
\noindent
\begin{minipage}[t]{0.50\linewidth}
    \vspace{0pt}
    Figure~\ref{fig:training_dynamics} shows that validation EH accuracy generally increases while verifiability-aware risk decreases.
    Both metrics fluctuate across checkpoints but show signs of leveling off after approximately 50 training steps.
    The multipliers initially increase during early budget violations and subsequently decline as the smoothed training risk falls below the budget.
    These updates depend on EMA-smoothed, pre-filtering training risk rather than the validation estimates shown in panel~(b).

    Figure~\ref{fig:lambda_initialization} presents exploratory LLaVA-OneVision-1.5-8B runs with $\lambda_0\in\{0,0.5p_H,p_H\}$ and $\delta=0.14$.
    All three trajectories initially rise and later decline,
    while retaining distinct multiplier magnitudes over the observed horizon.
\end{minipage}\hfill
\begin{minipage}[t]{0.46\linewidth}
    \vspace{0pt}
    \centering
    \captionsetup{type=figure}
    \includegraphics[width=\linewidth]{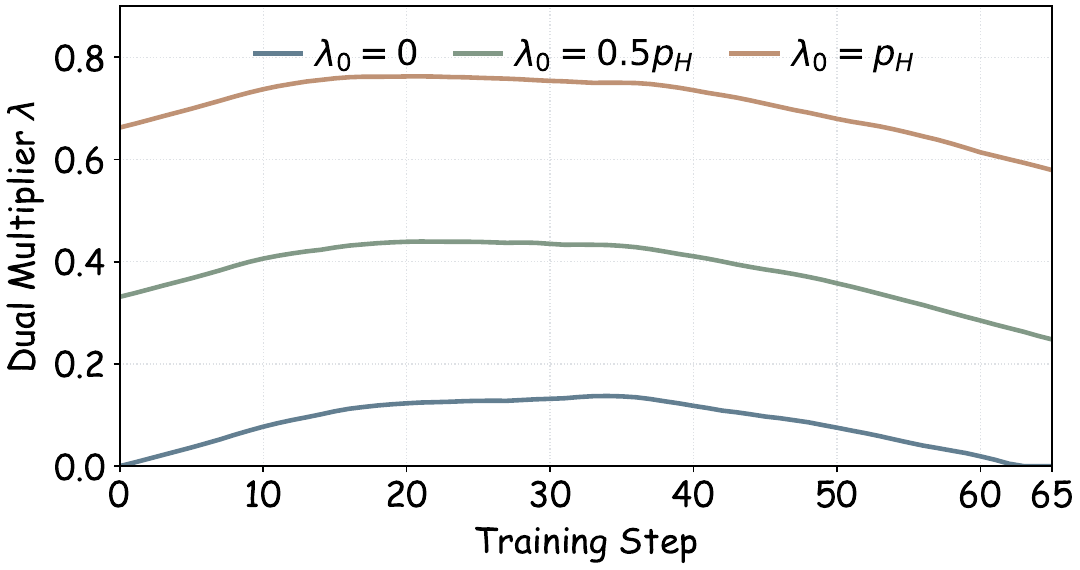}
    \caption{Exploratory dual-multiplier trajectories under different initializations on LLaVA-OneVision-1.5-8B,
    with $\delta=0.14$.}
    \label{fig:lambda_initialization}
\end{minipage}

\FloatBarrier
\Needspace{6\baselineskip}
\section{Human Evaluation Protocol}
\label{app:human_evaluation_protocol}

\subsection{Evaluation Sample Construction}
We used LLaVA-OneVision-1.5-8B as the base model for Base, DAPO, and DAPO-V. 
For each image, all three models generated a detailed description using the same prompt, provided in Appendix~\ref{app:human_study_prompt}. 
We manually inspected the descriptions against the corresponding images and extracted sentences containing visually unsupported or contradictory claims.
The resulting evaluation set comprised 100 hallucinated image--sentence pairs per model. 
We included correct image--sentence pairs as control items, amounting to 10\% of the hallucinated-item count.
These controls were not counted among the hallucinated items and were intended to discourage indiscriminate inconsistency judgments.

\subsection{Participants and Procedure}
Ten participants with backgrounds comparable to those in the dataset-construction study completed the evaluation.
Each trial presented an image and a selected sentence for consistency assessment.
Participants judged whether the sentence was consistent or inconsistent with the image, without knowing the model source.
Both metrics are computed exclusively on hallucinated image--sentence pairs.
Detection rate is the proportion of judgments correctly identifying these pairs as inconsistent.
Mean response time includes judgments resulting in both correct detections and missed hallucinations.
Visually consistent control items are excluded from both metrics.

\subsection{Prompt for the Human Evaluation}
\label{app:human_study_prompt}

\begin{figure}[!htbp]
    \centering
    \includegraphics[width=0.62\linewidth]{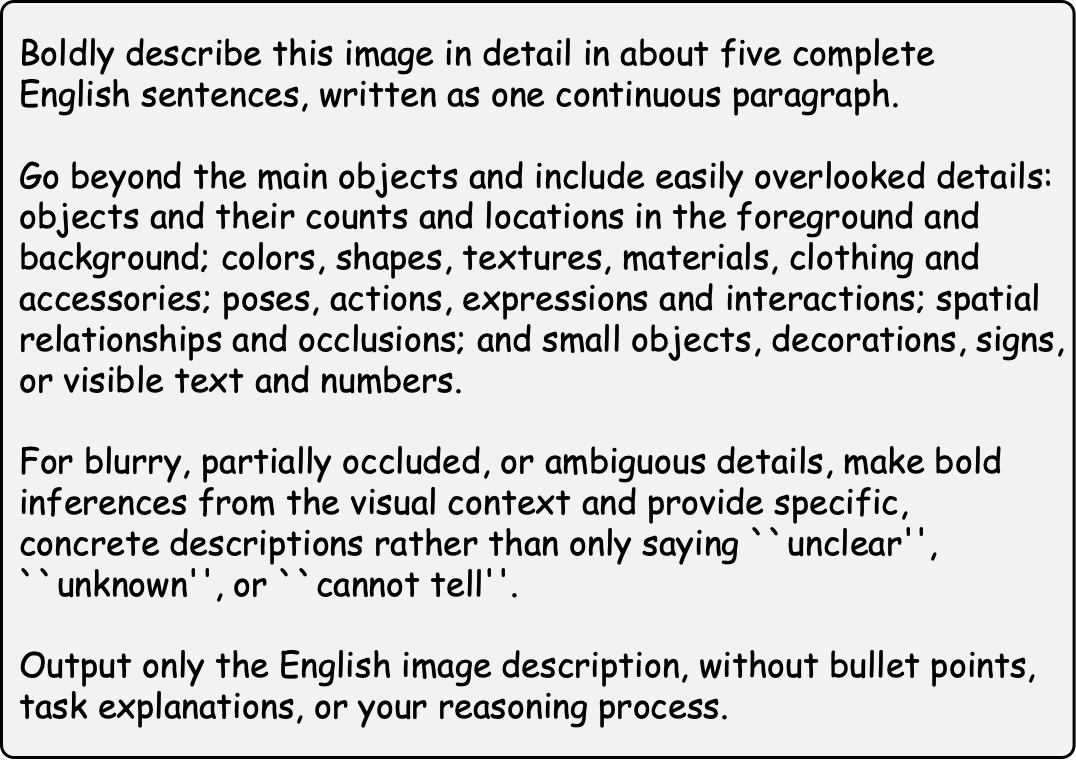}
    \caption{Image-description generation prompt used for human evaluation.}
    \label{fig:human_study_prompt}
\end{figure}

We used the same prompt, shown in Figure~\ref{fig:human_study_prompt}, to generate image descriptions with Base, DAPO, and DAPO-V for the human evaluation.

The prompt encourages the model to make specific inferences about ambiguous visual details, thereby inducing hallucinations to facilitate experimental data collection.

\end{document}